\documentclass[letterpaper]{article} 
\usepackage{aaai24}  
\usepackage{times}  
\usepackage{helvet}  
\usepackage{courier}  
\usepackage[hyphens]{url}  
\usepackage{graphicx} 
\usepackage{natbib}  
\usepackage{caption} 
\usepackage{algorithm}
\usepackage{algorithmic}
\usepackage{multirow}
\usepackage{siunitx}
\usepackage{amsmath}
\usepackage{booktabs}
\usepackage{bm}

\usepackage{newfloat}
\usepackage{listings}
\DeclareCaptionStyle{ruled}{labelfont=normalfont,labelsep=colon,strut=off} 
\floatstyle{ruled}
\newfloat{listing}{tb}{lst}{}
\floatname{listing}{Listing}
\title{Learning Multimodal Volumetric Features for Large-Scale Neuron Tracing}
\author {
    Qihua Chen,
    Xuejin Chen\thanks{ Corresponding author},
    Chenxuan Wang, 
    Yixiong Liu, 
    Zhiwei Xiong, 
    Feng Wu
}
\affiliations {
    National Engineering Laboratory for Brain-inspired Intelligence Technology and Application, University of Science and Technology of China, Hefei 230027, China.\\
    Institute of Artificial Intelligence, Hefei Comprehensive National Science Center, Hefei 230088, China.\\
    cqh@mail.ustc.edu.cn, xjchen99@ustc.edu.cn, \{wangcx01, yxliu99\}@mail.ustc.edu.cn,\\ \{zwxiong, fengwu\}@ustc.edu.cn
}

\usepackage{bibentry}

\begin{document}

\maketitle

\begin{abstract}
The current neuron reconstruction pipeline for electron microscopy (EM) data usually includes automatic image segmentation followed by extensive human expert proofreading. In this work, we aim to reduce human workload by predicting connectivity between over-segmented neuron pieces, taking both microscopy image and 3D morphology features into account, similar to human proofreading workflow. 
To this end, we first construct a dataset, named FlyTracing, that contains millions of pairwise connections of segments expanding the whole fly brain, which is three orders of magnitude larger than existing datasets for neuron segment connection. 
To learn sophisticated biological imaging features from the connectivity annotations, we propose a novel connectivity-aware contrastive learning method to generate dense volumetric EM image embedding. The learned embeddings can be easily incorporated with any point or voxel-based morphological representations for automatic neuron tracing.
Extensive comparisons of different combination schemes of image and morphological representation in identifying split errors across the whole fly brain demonstrate the superiority of the proposed approach, especially for the locations that contain severe imaging artifacts, such as section missing and misalignment. The dataset and code are available at \textcolor{blue}{https://github.com/Levishery/Flywire-Neuron-Tracing}.
\end{abstract}

\section{Introduction}
\label{sec:introduction}

\begin{figure}[htb]
    \centering
    \includegraphics[width=0.48\textwidth]{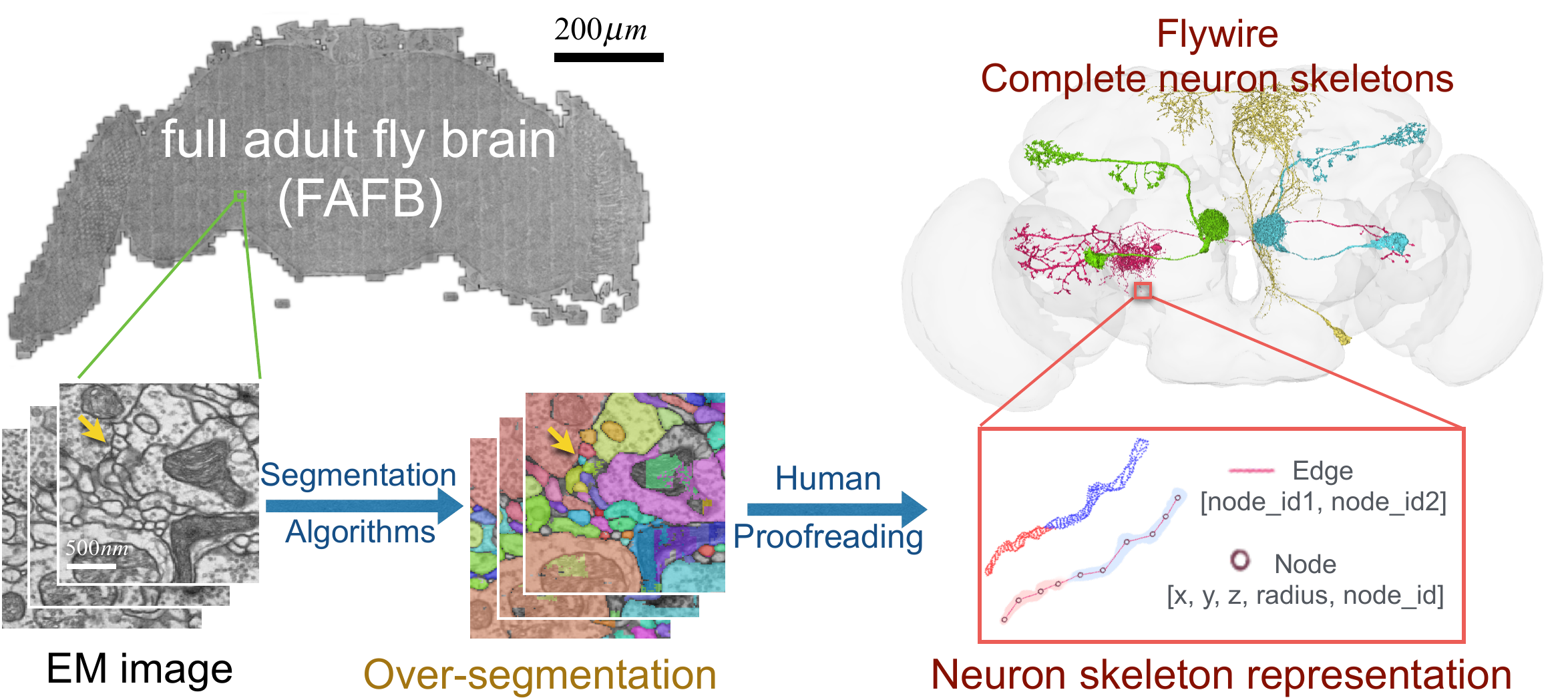}
    \caption{The pipeline of large-scale neuron reconstruction consists of EM image over-segmentation and human proofreading for complete neuron reconstruction. Each neuron is represented by a tree-structure skeleton. 
  }
    \label{fig:neuronal-data}
\end{figure}

The development of various imaging techniques, such as electron microscopy (EM), optical microscopy (OM), and Magnetic resonance imaging (MRI) scans, bring massive volumetric imaging data in biological and medical research fields. 
Segmenting 3D objects with complex structures and wide expansion in these high-resolution image volumes is non-trivial. 
Among various 3D segmentation tasks, connectomics, which aims to reconstruct and interpret neural circuits at the synaptic resolution, presents extreme challenges due to the huge volume of EM images and the significant morphological complexity of neurons. 
Notably, with the advent of advanced serial section electron microscopy imaging techniques, several EM data of $mm^3$-scale volumes have been published for the connectomics community, with the data sizes ranging from terabyte to perabyte~\cite{FAFB, h01, microns2021functional}.
Many automated volumetric image segmentation methods, either two-stage ~\cite{lee2021learning,beier2017multicut,funke2018large,sheridan2022local, lee2017superhuman} or end-to-end~\cite{FFN}, have been developed for neuron reconstruction.
While only exploiting the pixel-level local context in limited regions at the nanometer resolution, these methods have difficulty segmenting complete neurons while a single neuron typically spans a cable length of over one thousand micrometers.
When applied to large-scale serial section EM volume, complete neuron reconstruction becomes even more challenging due to various imaging artifacts, such as section missing and misalignment. Since a few merge errors could lead to several incorrectly merged neuronal processes and correcting the split errors is more straightforward, existing methods typically follow the consensus of over-segmentation where a neuronal process is segmented into many small fragments \cite{fafb-ffn,matejek2019biologically,FFN}. For example, a single neuron from a fly brain~\cite{FAFB} is usually segmented into hundreds of fragments by Flood-filling
networks (FFN)~\cite{fafb-ffn}. 
To reconstruct more complete neurons, intensive human proofreading efforts are required in current neuron tracing pipelines~\cite{dorkenwald2022flywire, h01}. Starting from the over-segmentation result, proofreaders trace complete neurons by inspecting the continuity in 3D morphology and image features between fragments and merging fragments when necessary, as Fig.~\ref{fig:neuronal-data} shows.  

A few methods have been developed to automatically detect and correct split and merge errors from the automatic segmentation~\cite{matejek2019biologically,zung2017error,VJain-MICCAI-2020}. 
\citet{matejek2019biologically} reduce split errors by extracting biologically-inspired graphs that indicate potential connectivity of over-segments using hand-designed geometric constraints. They train a 3D convolutional neural network (CNN) to predict the probability of merging based on segment morphologies. However, their method is trained and tested on small densely annotated blocks and not well generalized to various brain regions. 
Moreover, morphology is insufficient for reliable connectivity prediction as human proofreaders have to frequently check the 3D morphology as well as EM images of two neighboring segments during long-range neuron tracing.

In this paper, we propose a novel approach for predicting connectivity between over-segmented neuron pieces by taking both microscopy images and 3D morphology features into account, following human proofreading principles. 
Long-range 3D morphology could provide rough global information, while sophisticated voxel-wise image features capture fine-grained evidence of continuity within adjacent local regions for reliable predictions.
However, learning effective representations for the two modalities and fusing them are non-trivial. 
We propose a novel connectivity-aware contrastive learning method to learn dense volumetric EM image embedding from large-scale segment connectivity annotations.  
Then we fuse the EM image features with the 3D volume or surface representations to conduct connectivity prediction. 
The contributions of this work include:
\begin{itemize}
  \item We introduce a dataset that contains millions of connecting segment pairs across the whole fly brain region. The dataset is three orders of magnitude larger than existing datasets for neuron segment connection. Our dataset is comprehensive and could facilitate the development of automatic neuron-tracing techniques.
  
  \item We propose a novel connectivity-aware contrastive learning method
to learn dense volumetric EM image embedding from sparse segment connectivity. The image embedding can be fused with any point- or voxel-based 3D representations for reliable split error correction.

  \item Different combination schemes of image and morphological representation are extensively evaluated for split error identification across the whole fly brain, demonstrating the superiority of our proposed approach.
\end{itemize}

\begin{figure*}
    \centering
    \includegraphics[width=\textwidth]{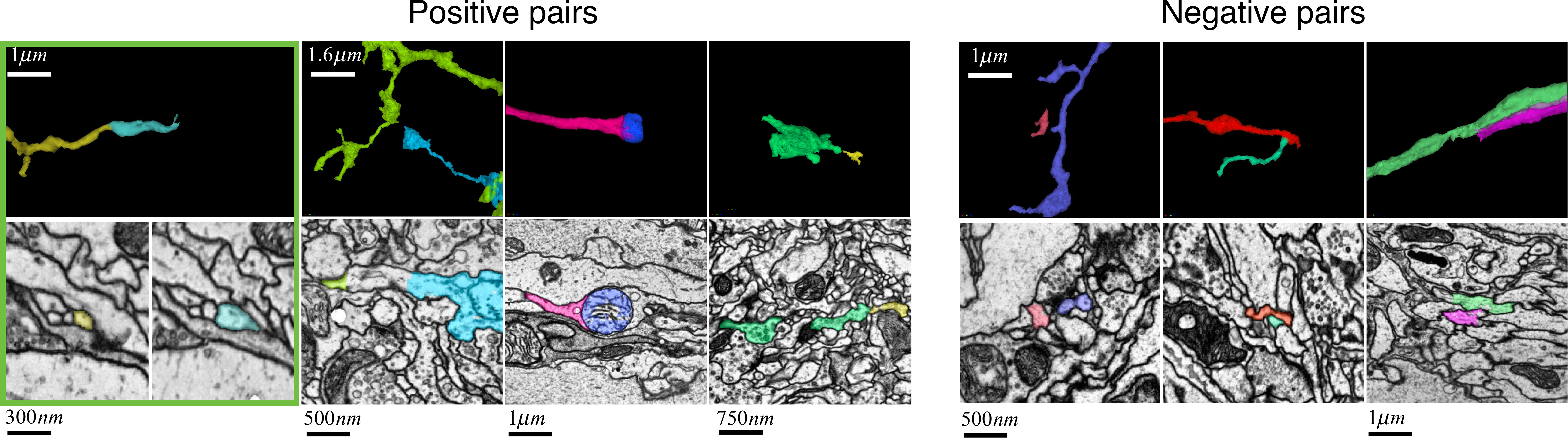}
    \caption{Example of sample pairs from FlyTracing. The first row shows their 3D meshes (segment instances denoted by color), with the representative EM image slices shown below. To determine whether two segments belong to the same neuron, human tracers frequently cross-reference their 3D meshes and adjacent EM image slices.}
    \label{dataset}
\end{figure*}
 \section{Related Work}
\label{sec:related-work}

 \paragraph{Voxel-based EM Neuron Segmentation.}
 Given a terabyte- or perabyte-scale EM image volume, many learning-based methods have been proposed for automatic segmentation. A popular series of methods train voxel-based convolutional neural networks to learn affinity maps~\cite{lee2021learning, sheridan2022local, funke2018large} and then apply non-parametric watershed transformation and agglomeration~\cite{beier2017multicut, wolf2018mutex}. 
 Flood-filling networks (FFN)~\cite{FFN} combine these two steps into one by gradually expanding segments from seed voxels. 
 Compared to affinity-based methods, FFN outperforms by a substantial margin, but at the expense of two orders of magnitude higher computational complexity. 
 In this work, we choose to correct the split errors from FFN segmentation results, instead of modifying the dense voxel-wise segmentation.

 \paragraph{Error Correction Methods.}

Although the afore-mentioned approaches determine most cell boundaries correctly, the remaining connection errors still require extensive human proofreading and correction. 
A few methods attempt to detect and correct errors automatically. \citet{zung2017error} firstly study error detection and correction for 3D neuron reconstruction. They train a 3D CNN to predict the map of split and merge errors and formulate error correction as an object mask pruning task. Other works focus on merge error correction using predefined neuron structural rules~\cite{VJain-MICCAI-2020, celii2023neurd}. 
\citet{matejek2019biologically} apply biological prior to
detect potential split errors, and then employ 3D CNNs to predict connection probability between the candidate pairs based on the segment morphology. We further explore the split error correction problem by introducing a comprehensive dataset and extensively studying different image and morphology representations for connectivity prediction. 

 \paragraph{Deep Metric Learning.}
The goal of metric learning is to learn a distance metric or embedding vectors such that similar samples are pulled closer and dissimilar samples are pushed away. Deep metric learning with pairwise~\cite{hadsell2006dimensionality} or triplet loss \cite{schroff2015facenet} performs favorably in image retrieval \cite{yang2018retrieving} and geo-localization \cite{shi2020optimal}. In the semantic and instance segmentation area, many algorithms employ metric learning to explore structural relations between pixels in the embedding space~\cite{zhou2022rethinking, lee2021learning, Wang_2021_ICCV}. The extracted pixelwise vectors of dense image embedding is significant helpful for downstream tasks. In this work, we also employ dense embeddings as a delicate image feature representation. The embedding vector of each voxel captures its neighboring appearance features, subsequently fused with point or voxel-based morphological representations to predict segment connectivity. Different from existing dense metric learning approaches, without requiring pixel-wise annotations, our method learns dense embeddings from sparse segment connectivity directly through a novel connectivity-aware contrastive loss.

\section{Task Formulation and Dataset}
\label{sec:datasets}
 
Fig. \ref{fig:neuronal-data} briefly illustrates the general pipeline of large-scale neuron reconstruction from EM image volume and various data used in our work.
The huge EM image volume is first over-segmented into a set of 3D segments using an off-the-shelf 3D segmentation approach. Subsequently, human proofreading is performed to obtain complete neurons that usually span several brain regions. The reconstructed neurons are commonly represented by tree structures composed of nodes and edges, either directly traced using annotation software such as CATMAID~\cite{saalfeld2009catmaid} or skeletonized from proofread segment surface using skeletonization algorithms such as TEASAR~\cite{sato2000teasar}. 
For each neuron, we register the over-segmented fragments with the neuron skeleton and obtain the connectivity relations between segments as the ground truth for training and testing.
 
In typical connectomics analysis workflows, human tracers start tracing from an interested neuron branch segment based on the over-segmentation results.
They identify the truncation point by examining its 3D surface mesh. Subsequently, they magnify the questionable area and determine which segment adjacent to the truncation point maintains neuronal continuity with the initial segment of interest, achieved by cross-referencing their 3D meshes and adjacent EM image slices.
For example, to trace the yellow branch in Fig.~\ref{dataset} (in the green box), a human tracer may zoom in to the terminal region, focusing on the lower-left image section. Then, the tracer transits to the adjacent section and sees that the area previously occupied by the yellow segment is now taken over by the blue segment, suggesting a potential connection between them. Typically, tracers proceed to verify their morphological continuity before merging.

To relieve the human proofreading workload for huge EM volumes, we propose to predict the connection probability of two segments, $S_a$ and $S_b$, considering their 3D morphology and the EM images from adjacent sections, mimicking the human tracing behavior. We learn the prediction function $f$ from a set of connected segment pairs ($f(S_a, S_b)=1$) that should be merged during proofreading and unconnected pairs ($f(S_a, S_b)=0$). 
 
\subsection{Dataset Construction}

\begin{table}[t]
    \caption{Overview of the datasets used in~\cite{matejek2019biologically} and our dataset \textbf{FlyTracing}. N/A denotes that the dataset is not public.
    }
    \centering
    \begin{tabular}{lS[table-format=1.1e2]lS[table-format=1.1e2]}
        \toprule
        \textbf{Dataset} & \textbf{Size ($\mu m^3$)} & \textbf{Method} & \textbf{\# Seg. Pairs} \\
        \midrule
        PNI & $\num{5.0e3}$ & Affinity & N/A \\
        Kasthuri & $\num{5.5e2}$ & Affinity & $\num{1.8e3}$ \\
        SNEMI3D & $\num{5.5e2}$ & Affinity & $\num{\sim e3}$ \\
        \textbf{FlyTracing} & $\num{3.2e6}$ & FFN & $\num{1.6e6}$ \\
        \bottomrule
    \end{tabular}
    \label{dataset-compare}
\end{table}

 \subsubsection{Dataset Overview.}
Available datasets for neural segment connection tasks mainly originated from densely annotated blocks, limited in scale and diversity. 
In contrast, we extract segment connectivity from the crowd-sourced proofreading results throughout an entire fly brain. 
As Table \ref{dataset-compare} shows, our dataset \textbf{FlyTracing} surpasses existing datasets by three orders of magnitude, regarding the volume size and number of connected segment pairs.
The source EM images for FlyTracing are from a complete adult Drosophila brain, imaged at $4\times4 nm$ resolution and sectioned with the thickness of $40 nm$, known as the “full adult fly brain” (FAFB) dataset~\cite{FAFB}. The image sections are first preprocessed through local re-alignment and irregular section substitution, and segmented through a multi-scale FFN segmentation pipeline~\cite{fafb-ffn}, referred to as FAFB-FFN1. The proofread neuron skeletons are supplied by FlyWire~\cite{dorkenwald2022flywire}. 
Despite the availability of affinity-based automatic segmentation from FlyWire, we choose FAFB-FFN1 due to its adherence to over-segmentation consensus, i.e., fewer merging errors than affinity-based segmentation results. 

 \subsubsection{Segment-Neuron Registration.} 
 To generate the ground truth of connectivities of EM segment pairs, we register the FFN segmentation results with the proofread neuron skeletons.
 We design an automatic EM segment-neuron registration method that can be applied to any large-scale connectomics datasets with proofread neuron skeletons and over-segmentation results. 
 With permission from Flywire, we obtain the surface meshes of the proofread neurons and skeletonize them using the skeletor tool~\cite{philipp_schlegel_2022_7308283}. 
 Since we focus on tracing neurites with tree-like structures, we cut off the cell body fibers from the neuron segments.

To register the massive over-segmented fragments with a human proofread neuron, given its neuron skeleton $T_{n}$, we associate each skeleton node with its nearest segment.
Since the extracted skeletons and EM image segmentation results inevitably contain errors and noises, a few nodes are occasionally assigned to segments that do not belong to the right neuron. 
To mitigate the assignment errors, we calculate the chamfer distance from the segment skeleton $T_{S}$ to the neuron skeleton $T_{neu}$, subsequently discarding the segments and their corresponding nodes whose $CD(T_{S}, T_{neu})>2\bar{r}$. 
Here, $\bar{r}$ denotes the estimated average radius of the local branch: $\bar{r} = \frac{1}{ |\Omega_S|} \sum_{i\in \Omega_S} r_i$, where $\Omega_S$ is the set of all the neuron skeleton nodes associated to segment $S$. 
Once every node is assigned with a corresponding segment, we traverse along the neuron skeleton and identify edges between nodes that are assigned to different segments as bridging edges:
\begin{equation}
    E_{bridge}(T_{neu}) = \{edge(\mathbf{v}_i, \mathbf{v}_j)| S_a \neq S_b\},
\end{equation}
where $\mathbf{v}_i$, $\mathbf{v}_j$ represent two adjacent nodes in $T_{neu}$, and $S_a,S_b$ are their corresponding segments, indicating $S_a$ and $S_b$ should be merged as the same neuron via $edge(\mathbf{v}_i,\mathbf{v}_j)$. 
\begin{figure}[t]
    \centering
    \includegraphics[width=0.48\textwidth]{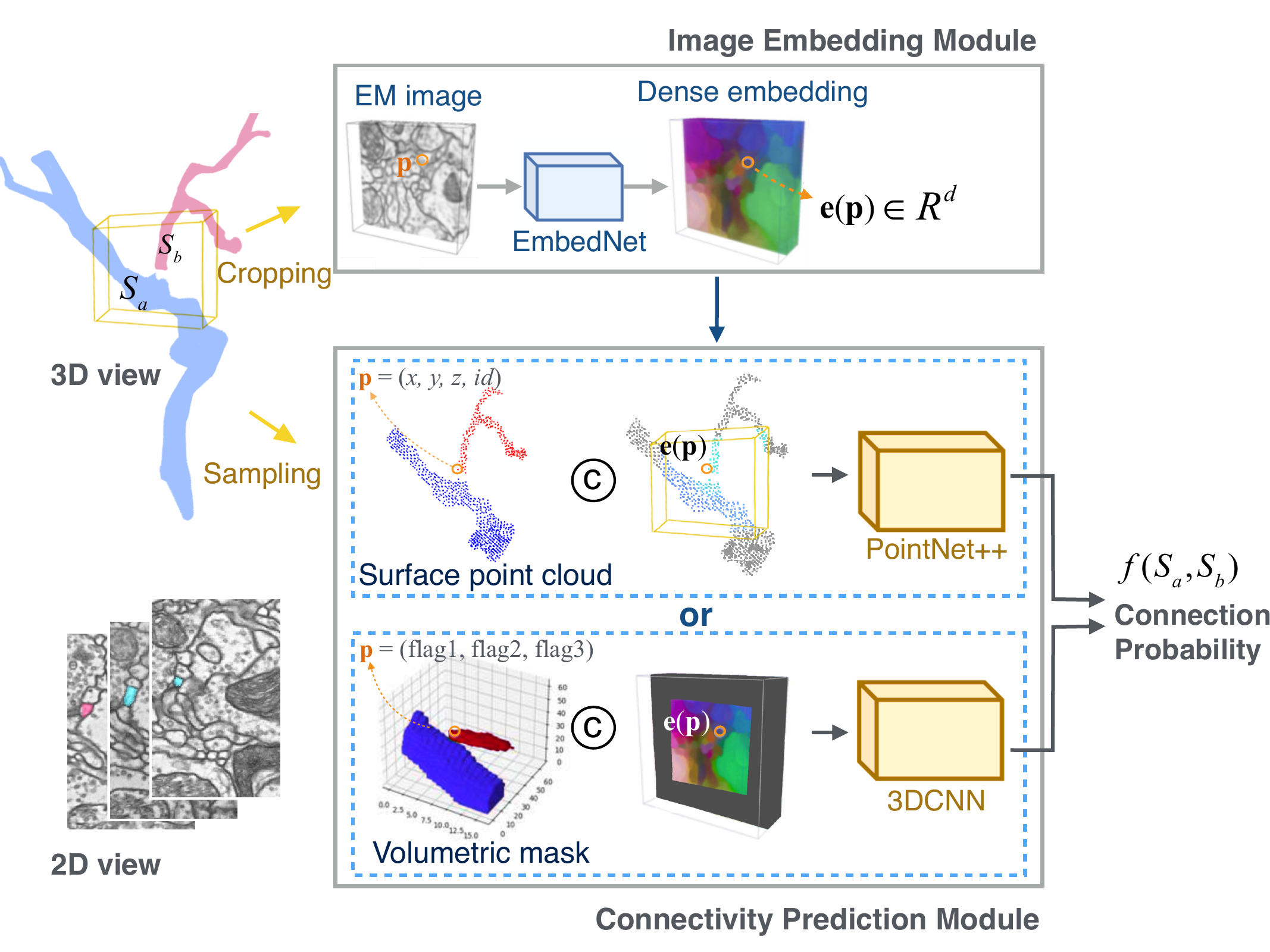}
    \caption{Our connectivity prediction framework fuses local volumetric image features extracted by the EmbedNet with 3D morphology, optionally represented by point cloud or volumetric masks.}%
    \label{fig:3Dmodel}
\end{figure}

Consequently, we collect the entire bridging edge set from $23,769$ proofread neurons.
The average count of bridging edges per neuron is 810. 
For each bridging edge $edge(\mathbf{v}_i, \mathbf{v}_j)$, assuming ${\mathbf{c}}_{i,j}$ is the midpoint of $\mathbf{v}_i$ and $\mathbf{v}_j$, we add random shift to the coordinate of ${\mathbf{c}}_{i,j}$ and obtain ${\hat{\mathbf{c}}}_{i,j}$ as the estimated truncated point. 
The segment pair $(S_a, S_b)$ is annotated as a positive connecting pair ($f(S_a, S_b)=1$).
Any segment $S_c$ that is located in the cube of size $H\times W\times D$ centered at ${\hat{\mathbf{c}}}_{i,j}$ and $S_c \neq S_b$ is labeled with $S_a$ as a negative pair ($f(S_a, S_c)=0$).

 \subsubsection{Training and Test Block Partition.}
The positive segment pairs across the entire fly brain are partitioned into blocks based on their location in the brain, each spanning a volume size of $26\times 26\times 1\mu m^3$. We select $4,000$ blocks, each of which contains a minimum of $350$ positive segment pairs for training and testing. 
 $1,000$ blocks are selected randomly as the training and validation set for the image embedding network and pairwise connectivity prediction models. The rest $3,000$ blocks are used for testing of connectivity prediction.

\section{Our Method}
\label{sec:method}

We predict the connection probability of a pair of segments by integrating image embedding with their 3D morphological features. We design a novel connectivity-aware contrastive learning method to learn dense image embedding from sparse segment connectivity annotations.
\begin{figure*}[t]
    \centering
    \includegraphics[width=0.88\textwidth]{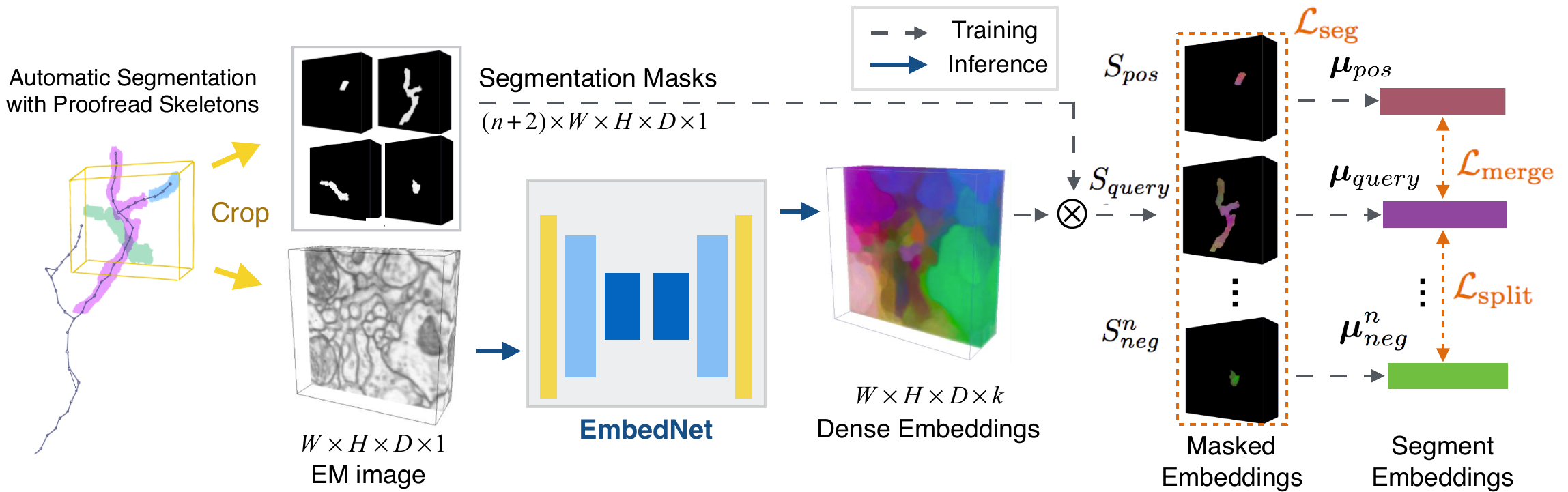}
    \caption{Our volumetric image embedding network. For a pair of adjacent segments, a small EM volume centered at the truncation point is cropped and fed into EmbedNet to extract per-voxel $k$-dimensional embeddings. The EmbedNet is trained via a contrastive loss based on the pairwise segment connectivity of the query segment and the other neighboring segments.}%
    \label{fig:embedding}
\end{figure*}

\subsubsection{3D Morphological Representation of Segments.}
The geometric property of segments provides rough clues for connection. For instance, certain patterns like L-shaped or arrow-shaped junctions are uncommon neuronal structures. 
There are several 3D representation forms for expressing segment morphology. For example, EdgeNetwork \cite{matejek2019biologically} uses volumetric masks for segments and applies 3D convolutional neural network (3DCNN) to encode the segment morphology. 
Each voxel carries three labels indicating whether it belongs to segments $S_a$, $S_b$, and $S_a \cup S_b$ respectively. Besides the 3D segment masks, we also explore the surface point cloud as a more efficient morphology representation. Unlike 3DCNNs that apply dense operation for both surface voxel and inner voxels of the target segments, point-based models only process sparse surface points. Centering at the truncation point ${\hat{\mathbf{c}}}_{i,j}$ of the segment to trace, we crop a 3D volume and extract the segment section contours on each image slice and collect all the contour points to compose a point cloud. 
Each point $\mathbf{p}=(x,y,z, id)$, where $(x,y,z)$ refer to the physical coordinate and $id\in\{0,1\}$ signifies surface points from whether $S_a$ or $S_b$. 
The point clouds are further downsampled by farthest point sampling (FPS) to a fixed number of points and normalized in a unit cube. We employ PointNet++ \cite{PointNet++-NIPS-2017} as the morphological feature encoder and connectivity prediction model for the point cloud inputs.

\subsubsection{Fusing Image Features with 3D Representations.}
The segment shape might be distorted due to image distortions (e.g. misalignment between adjacent EM sections) and segmentation errors. Consequently, human tracers rely on the original EM images as the decisive evidence. Sometimes the evidence could lie on just a few voxels. Therefore, we design an image embedding network called EmbedNet to generate delicate and discriminative features for every voxel. From a local EM image cube, our EmbedNet extracts a $k$-dimensional image feature vector for each voxel. To predict the segment connectivity, the image features are integrated with the point geometric features or voxel masks optionally, as illustrated in Fig. \ref{fig:3Dmodel}. 
For the volume representation, the segment masks and image embedding vectors are concatenated directly voxel by voxel. 
Regarding the surface point cloud, we use the average embedding vectors in a $7 \times 7 \times 3$ neighborhood of each point $\mathbf{p}$ as its local image embedding $\mathbf{e}(\mathbf{p})$.
The $k$-D average embedding feature is then concatenated to $(x,y,z,id)$ of each surface point.

\subsubsection{Learning Volumetric Image Embedding.} 
\label{Sec:learn-embedding}
To learn the image feature representations, we follow the typical supervised contrastive learning framework~\cite{lee2021learning,de2017semantic,zhou2022rethinking} with a contrastive loss that encourages close embeddings between connected segments and distinct embeddings of segments from different neurons.
Fig.~\ref{fig:embedding} shows the pipeline of our EM image embedding learning. 
Existing dense embedding learning methods for segmentation usually apply a loss that mainly consists of two parts: intra-cluster variance that measures the deviation of the embeddings of all segment voxels from the cluster center and inter-cluster distance that forces distinct clusters.
In comparison, we focus more on segment-level connectivity rather than voxel-wise segmentation.  
Consequently, we design a novel segment connectivity-oriented contrastive loss to enforce the similarity of the mean embeddings between segments belonging to the same neuron.

Our connectivity-oriented contrastive loss consists of a merge loss for positive pairs and a split loss for negative pairs. 
As shown in Fig.~\ref{fig:embedding}, assuming we have a positive connection pair that contains a query segment $S_{query}$ and a positive segment $S_{pos}$ from the same neuron, we crop an EM cube with the over-segmentation masks. Then $n$ segments $\{S_{neg}^1,...,S_{neg}^n\}$ are randomly sampled from adjacent areas to form negative connection pairs with $S_{query}$ and $S_{pos}$ respectively. 
As the EmbedNet extracting per-voxel image embedding feature $\mathbf{e}$, for each segment that contains $M$ voxels within the cropped cube, we compute the mean embedding feature as the segment embedding: $\bm{\mu} = \frac{1}{M} \sum^{M}_{i=1} \mathbf{e}_i$. 
Then the merge loss and split loss are computed by:
\begin{equation}
    \begin{aligned}
     \mathcal{L}_{\text {merge }} &=\left\|\bm{\mu}_{query}-\bm{\mu}_{pos}\right\|^2, \\
     \mathcal{L}_{\text {split}} &=\frac{1}{n}\sum_{i=1}^n\max \left(2 \delta_d-\left\|\bm{\mu}_{query}-\bm{\mu}_{neg}^i\right\|, 0\right)^2 \\
    & + \frac{1}{n} \sum_{i=1}^n\max \left(2 \delta_d-\left\|\bm{\mu}_{pos}-\bm{\mu}_{neg}^i\right\|, 0\right)^2.
    \end{aligned}
\end{equation}

This design enables our EmbedNet to make use of the large-scale pairwise connection dataset and learn robust features against diverse imaging quality. 
However, the learning of dense embedding is difficult to converge with only sparse segment-level connectivity supervision. Therefore, we following existing segmentation-based embedding methods to further encourage the embedding to form compact clusters within each segment. 
Without dense segmentation annotations for each complete neuron, which are expensive to manually label for a large-scale dataset with diverse image characteristics, we use the automatic over-segmentation results of EM images as pseudo-segmentation masks and adopt the segmentation clustering loss defined in \cite{lee2021learning}:
\begin{equation}
\label{eq:segmentation-loss}
    \mathcal{L}_{seg} = \mathcal{L}_{\mathrm{int}} + \mathcal{L}_{\mathrm{ext}} + \gamma\mathcal{L}_{\mathrm{reg}},
\end{equation}
where $\gamma=0.001$.

The overall loss is composed of the segment connectivity contrastive loss and the voxel segmentation clustering loss:
\begin{equation}\label{eq:overall-loss}
        \mathcal{L}_{\text {total }} = \lambda_1  \mathcal{L}_{\text {merge }} + \lambda_2 \mathcal{L}_{\text {split}} + \lambda_3 \mathcal{L}_{\text {seg}}.
\end{equation}
During training, we use $\lambda_1=0.1$, $\lambda_2=1$ and $\delta_d=1.5$. Since $\mathcal{L}_{\text {seg}}$ is contradictory with $\mathcal{L}_{\text {merge}}$, we initialize $\lambda_3$ to $1$ and linearly decrease it to $0.2$.

\begin{table*}[t]
\renewcommand{\arraystretch}{1.05}
 \caption{Performance comparison on our FlyTracing test blocks.}
  \centering
  \begin{tabular}{|l|c@{\hspace{7pt}}c@{\hspace{8pt}}c|c@{\hspace{8pt}}c@{\hspace{8pt}}c|c@{\hspace{8pt}}c@{\hspace{8pt}}c|c@{\hspace{8pt}}c@{\hspace{8pt}}c|}
    \hline
    Dataset & \multicolumn{3}{c|}{Avg. on 3K test blocks} & \multicolumn{3}{c|}{Misalignment }& \multicolumn{3}{c|}{Missing sections } & \multicolumn{3}{c|}{Mixed degradation}\\
    \hline
    Method   & Rec. & Prec. & F1 & Rec. & Prec. & F1 & Rec. & Prec. & F1 & Rec. & Prec. & F1\\
    \hline
    Connect-Embed o. &  $0.890$ & $0.926$ & $0.908$ & $0.738$ & $0.886$ & $0.805$ & $0.890$ & $0.937$ & $0.914$ & $0.810$ & $0.907$ & $0.856$\\
    \hline
    EdgeNetwork  & $0.859$ & $0.969$ & $0.911$ & $0.626$ & $0.944$ & $0.752$& $0.690$ & $0.958$ & $0.802$& $0.699$ & $0.960$ & $0.809$\\
    + Intensity & $0.881$ & $0.970$ & $0.923$ & $0.658$ & $0.946$ & $0.776$& $0.748$ & $0.952$ & $0.838$& $0.713$ & $0.960$ & $0.818$ \\
    + Seg-Embed & $0.870$ & $0.969$ & $0.916$& $0.668$ & $0.937$ & $0.782$& $0.766$ & $0.945$ & $0.846$& $0.749$ & $0.955$ & $0.840$\\
    + Connect-Embed & $0.883$ & $0.971$ & $0.925$& $0.673$ & $0.937$ & $0.783$& $0.748$ & $0.960$ & $0.841$& $0.743$ & $0.962$ & $0.839$\\
    \hline
    PointNet++  & $0.847$ & $0.939$ & $0.891$& $0.628$ & $0.893$ & $0.737$ & $0.693$ & $0.895$ & $0.781$& $0.672$ & $0.903$ & $0.771$\\
    + Intensity  & $0.865$ & $0.949$ & $0.905$ & $ 0.646$ & $0.911$ & $0.756$ & $0.767$ & $0.892$ & $0.825$ & $0.748$ & $0.914$ & $0.823$\\
    + Seg-Embed & $0.914$ & $0.954$ & $0.934$ & $0.739$ & $0.930$ & $0.824$& $0.812$ & $0.924$ & $0.864$& $0.816$ & $0.914$ & $0.862$\\
    + Connect-Embed & $\mathbf{0.932}$ & $\mathbf{0.972}$ & $\mathbf{0.952}$& $\mathbf{0.834}$ & $0.941$ & $\mathbf{0.884}$ & $\mathbf{0.909}$ & $\mathbf{0.963}$ & $\mathbf{0.935}$ & $\mathbf{0.883}$ & $0.949$ & $\mathbf{0.915}$\\
    
    \hline
  \end{tabular}
  \label{over-all}
\end{table*}

\section{Experiments and Results}
\label{Experiment}
 
\subsubsection{Model Configurations.} 
Our EmbedNet follows the architecture of residual symmetric U-Net~\cite{lee2017superhuman}, which tackles anisotropic serial-sectioning EM images with mixed 2D and 3D convolutional layers. We add a Squeeze-and-Excitation block~\cite{hu2018squeeze} to each scale to enhance the embedding expression capability. We set the input volume size as $129 \times 129 \times 17$ under the voxel resolution of $16\times16\times40 nm^3$, and the embedding dimension $k=16$. 
The EdgeNetwork~\cite{matejek2019biologically} employs a 3DCNN composed of three convolutional layers with filter sizes 16, 32, and 64, respectively, taking a cube with a side length of $1200nm$ and resized to $52\times52\times18$ as the input. The point clouds are sampled within a cube in size of $2560nm$, and down-sampled to $2048$ points by FPS.

\subsubsection{Training Details.}
We train the embedding network with AdamW optimizer for $500k$ iterations with a batch size of $8$, and apply data augmentation including random rotation, rescale, flip, and grayscale intensity augmentation. The initial learning rate is set to $0.002$ with warmup and a step decay scheduler. 
The number of negative sample pairs $n=20$. 
To enhance the learning from EM images with section missing and misalignment, we further fine-tune our EmbedNet on a subset of training blocks where the embedding network has a higher loss at the first round of training.  

The EdgeNetworks (with and without image embedding features) are trained with AdamW optimizer for $45k$ steps with a batch size of $128$. The PointNet++ models are trained with AdamW optimizer for $200k$ steps with a batch size of $92$. 
Because in real neuron tracing scenarios, there are much more adjacent segment pairs that do not belong to the same neuron, we train our connectivity prediction network with more negative segment pairs by setting the ratio between positive samples and negative samples to $3:7$. 
All the $422k$ positive segment pairs in the 1000 training data blocks are included during training and validation.

\subsection{Results of Segment Connectivity Prediction}

\subsubsection{Metrics.} 
To evaluate the segment connectivity prediction performance, we use the recall and precision metrics. On the $3000$ testing blocks, all the $1,178k$ positive segment pairs are used for evaluation. For each positive pair, we pick one segment as the query segment of the pair and randomly sample one of its neighboring segments from another neuron near the truncation point to form a negative pair. 
Two segments $S_a, S_b$ are determined as connected as the same neuron if $f(S_a,S_b)>0.5$.

\subsubsection{Comparison of Multiple Methods.}
We evaluate various combinations of image and morphology representations and report the results in Table~\ref{over-all}.
For the image features, we also evaluate `Intensity' and `Seg-Embed'. `Intensity' corresponds to the raw voxel intensity of the EM image, while `Seg-Embed' refers to the image embedding approach for neuron segmentation proposed in \cite{lee2021learning}. 
In our experiment, the `Seg-Embed' embedding is trained using dense neuron segmentation annotation from the CREMI dataset, sourced from three sub-volumes of FAFB. Notably, the embedding in \cite{lee2021learning} serves as a surrogate of the affinity map for neuron segmentation, rather than for correcting segmentation split errors. 
Image features are not exploited in the vanilla `EdgeNetwork' or `PointNet++' model.
In addition to these combinations, we also present the prediction performance using single modalities. `Connect-Embed o.' denotes that we determine $f(S_a,S_b)=1$ if the distance between the mean embedding of the two segments $\left\|\bm{\mu}_{query}-\bm{\mu}_{candidate}\right\|<\delta_d$. 

\begin{figure}[t]
    \centering
    \includegraphics[width=0.45\textwidth]{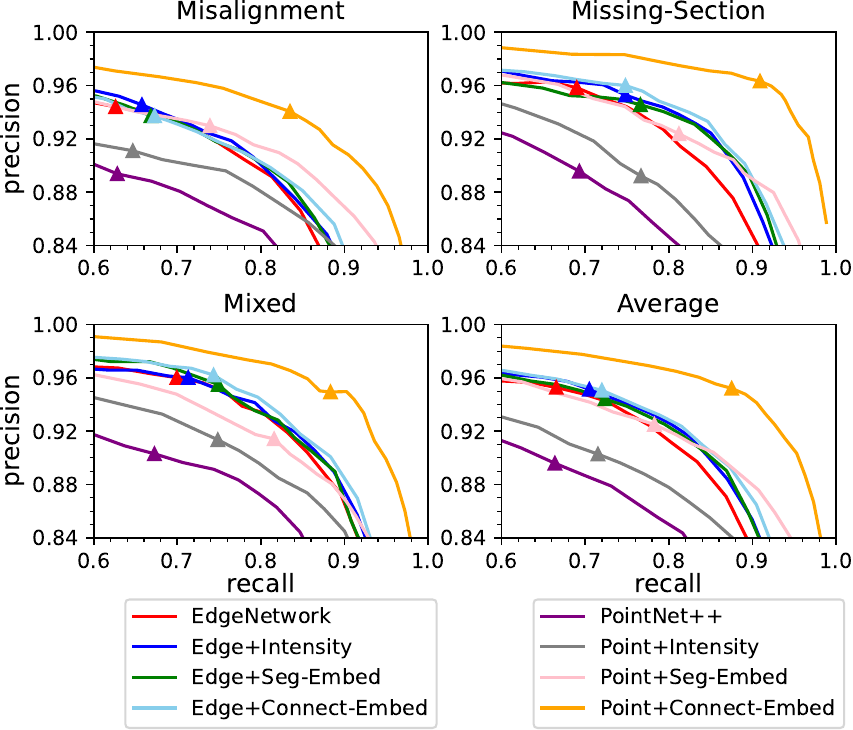}
    \caption{Precision-recall curves of different models on challenging blocks with image degradation. The triangle markers denote the performance using threshold $f(S_a,S_b)>0.5$. } 
    \label{thresh}
\end{figure}

As Table~\ref{over-all} shows, the combination of point cloud representation and our connectivity-aware embedding ``+Connect-Embed' performs the best. 
Compared with 3D voxel-based model EdgeNetwork, PointNet++ reaps greater advantages from incorporating image features. 
Since 3DCNNs suffer cubical computation increases when enlarging the input volume size, the EdgeNetworks have to consider the trade-off between input resolution and field of view, preventing them from getting the advantages of local fine-grained image features and global morphological features simultaneously. 
On the contrary, our proposed fusion of point cloud with image embedding effectively preserves local sophisticated image characteristics, while encompassing morphology information across a broad field of view within a fixed number of input points. 
In addition, among the three manners of image feature encoding, including intensity only, segmentation-based embedding, and our connectivity-aware embedding, our embedding is the most effective for segment connectivity prediction. %

Particularly for the EM volume blocks that suffer from severe image degradations, such as misalignment or section missing, the proposed PointNet++ (+Connect-Embed) yields remarkable enhancements to competitors. 
Among the $3,000$ test blocks, we select $10$ blocks containing $300\!\sim\!800nm$ misalignment, $10$ blocks containing $2\!\sim\!3$ missing sections, and $5$ blocks suffering from both degradations (typically one section missing followed by misalignment of $200\!\sim\!300nm$). 
Fig.~\ref{thresh} depicts the precision-recall curves of the models on the low-quality EM blocks.
Notably, the curve of our proposed feature combination consistently outperforms the others by a substantial margin.
This significant improvement is mainly brought by the proposed embeddings that can compensate for the information loss of segment morphology at the regions with image degradation and shape distortion. The raw voxel intensity retains only sparse low-level information from the image when combined with the point cloud, therefore providing limited improvement. On the other hand, without the supervision of pair-wise contrast of segments throughout the brain, Seg-Embed \cite{lee2021learning} fails to generate robust embeddings against imaging artifacts. In contrast, our embedding approach provides both structural and fine-grained image features.

\begin{figure}[t]
    \centering
    \includegraphics[width=0.48\textwidth]{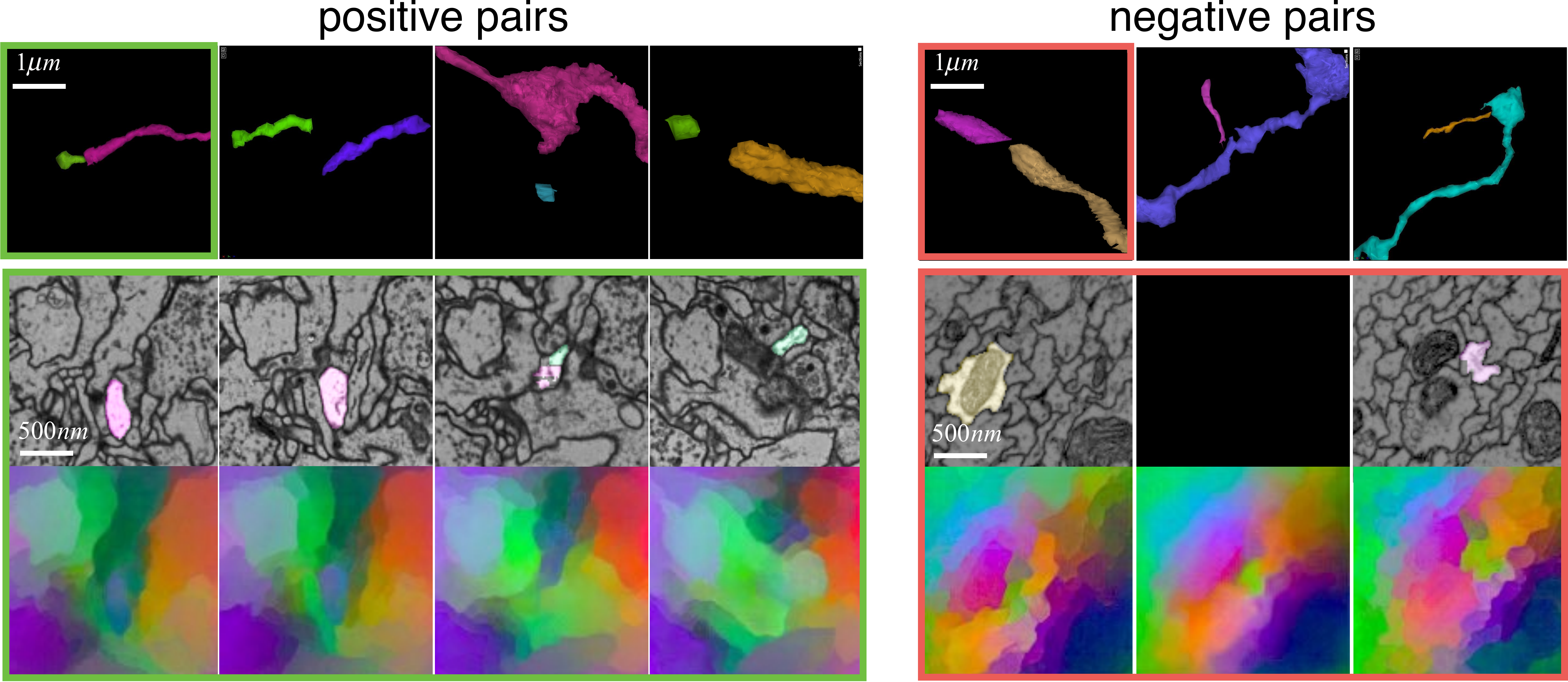}
    \caption{Visualization of several segment pairs whose connection relations are correctly predicted by model PointNet++ with `Connect-Embed'. In the two lower rows, we show the representative EM image sections and corresponding embeddings of the samples in the color boxes. }
    \label{visual}
\end{figure}

\subsubsection{Embedding Visualization.}
 In Fig.~\ref{visual}, we show two sets of segment pairs whose connection relationships are correctly classified by our PointNet++ with `Connect-Embed' model. 
We visualize the embeddings using PCA to project the high-dimensional embedding vector onto the RGB color space. 
Without the image embedding, the PointNet++ model fails to predict the connectivity of these segment pairs because of the ambiguous morphology. The positive pairs are often subject to image degradation, resulting in morphological distortions or too small segments without sufficient evidence for connection. 
On the other hand, the morphology of misclassified negative pairs is hard to distinguish from positive connectivity pairs. 
In contrast, our embedding provides distinctive clues for neuron tracing, effectively complementing the distorted morphological features. Despite the variations in the location and appearance of the neurons across sections, our embedding remains discriminative and consistent. 

\subsubsection{Ablation Study of EmbedNet.}
We measure the discriminative ability of EmbedNet under different loss configurations. The discriminative ability is defined as the rank of $\left\|\bm{\mu}_{query}\!-\!\bm{\mu}_{pos}\right\|$ among all $n+1$ candidates $\{\!\left\|\bm{\mu}_{query}\!-\!\bm{\mu}_{pos}\right\|\!,\left\|\bm{\mu}_{query}\!-\!\bm{\mu}_{neg}^1\right\|\!,...,\left\|\bm{\mu}_{query}\!-\!\bm{\mu}_{neg}^n\right\|\!\}$. 
We report the mean rank on $3$ test blocks. The adaptive weighting strategy performs the best since $L_{seg}$ gives dense supervision of voxel-wise segmentation in earlier steps, while decreasing the weight enforces the model to focus on learning discriminative features from pairwise connection annotations, i.e., $L_{split}$ and $L_{merge}$, in later steps.

\begin{table}[t] 
\setlength\tabcolsep{6pt}
\centering
\caption{Discriminative ability (the lower the better) of the embedding network with different weights of $L_{seg}$ ($\lambda_3$), where $\infty$ indicates that the model is trained with $L_{seg}$ only, and `Adaptive' denotes the decreasing strategy of $\lambda_3$. }
\label{ablation}
\begin{tabular}{l|ccc|c}
\hline
$\lambda_3$ & Block A & Block B & Block C & Average \\
 \hline
$\infty$ & $4.49$ & $4.18$ & $3.52$ & $4.07$ \\
$1$ & $2.46$ & $2.14$ & $1.85$ & $2.15$ \\
$0.1$ & $1.73$ & $1.60$ & $1.36$ & $1.57$ \\
$0$ & $2.34$ & $2.07$ & $2.04$ &$2.15$ \\
\hline
Adaptive & $\mathbf{1.59}$ & $\mathbf{1.45}$ & $\mathbf{1.19}$ & $\mathbf{1.41}$ \\
\hline
\end{tabular}
\end{table}
 
\subsubsection{Tracing Result on FAFB Mushroom Body.}
We also evaluate our proposed model PointNet++ with the `Connect-Embed ' within a real tracing scenario on the FAFB dataset. Following the evaluation setting described in \cite{fafb-ffn}, we evaluate the segmentation with a set of 166 ground truth neuronal skeletons produced by human tracers \cite{FAFB} using modified expected run length (ERL) metric from \cite{wei2021axonem}. 
We crop a volume of size $64\times44\times56$ $\mu m^3$ from the mushroom body region which contains around $10\%$ of the ground truth skeleton nodes and exclude the subvolumes in the training set. The candidate pairs and truncated points are estimated from the endpoints of segment skeletons as \cite{matejek2019biologically}. Using FAFB-FFN1 as the initial EM image segmentation, with a merging threshold of $0.98$, our automatic proofreading with pairwise segment connectivity prediction increases ERL by $28.8\mu m$ (a relative increase of $8.1\%$ ).

\section{Conclusion}
We explore automatic split error correction in the neuron reconstruction pipeline by introducing a comprehensive dataset and extensively studying different image and morphology representations for connectivity prediction. Additionally, we propose a novel connectivity-aware contrastive learning method to learn discriminative EM image embeddings. Integrating the image embeddings with point clouds shows superiority over other methods in identifying connectivity between over-segmented neuron fragments in a huge volume of electronic microscopy images of a fly brain. For the future work, we intend to study beyond pairwise connection prediction. Reasoning the segment connectivity by considering multiple candidates together and the tracing history could yield more accurate decision and facilitate fully automatic complete neuron tracing.

\section*{Acknowledgments}
This work was supported by the National Natural Science Foundation of China under Grant 62076230 and the Fundamental Research Funds for the Central Universities under Grant WK3490000008. 
We thank the Princeton FlyWire team and members of the Murthy and Seung labs, as well as members of the Allen Institute for Brain Science, for the development and maintenance of FlyWire (supported by BRAIN Initiative grants MH117815 and NS126935 to Murthy and Seung). We also acknowledge members of the Princeton FlyWire team and the FlyWire consortium for neuron proofreading and annotation.

\bibliography{aaai24}

\begin{thebibliography}{29}
\providecommand{\natexlab}[1]{#1}

\bibitem[{Beier et~al.(2017)Beier, Pape, Rahaman, Prange, Berg, Bock, Cardona, Knott, Plaza, Scheffer et~al.}]{beier2017multicut}
Beier, T.; Pape, C.; Rahaman, N.; Prange, T.; Berg, S.; Bock, D.~D.; Cardona, A.; Knott, G.~W.; Plaza, S.~M.; Scheffer, L.~K.; et~al. 2017.
\newblock Multicut brings automated neurite segmentation closer to human performance.
\newblock \emph{Nature methods}, 14(2): 101--102.

\bibitem[{Brabandere, Neven, and Van~Gool(2017)}]{de2017semantic}
Brabandere, B.~D.; Neven, D.; and Van~Gool, L. 2017.
\newblock Semantic Instance Segmentation for Autonomous Driving.
\newblock In \emph{Proceedings of the IEEE Conference on Computer Vision and Pattern Recognition (CVPR) Workshops}.

\bibitem[{Celii et~al.(2023)Celii, Papadopoulos, Ding, Fahey, Wang, Papadopoulos, Kunin, Patel, Bae, Bodor et~al.}]{celii2023neurd}
Celii, B.; Papadopoulos, S.; Ding, Z.; Fahey, P.~G.; Wang, E.; Papadopoulos, C.; Kunin, A.~B.; Patel, S.; Bae, J.~A.; Bodor, A.~L.; et~al. 2023.
\newblock NEURD: automated proofreading and feature extraction for connectomics.
\newblock \emph{BioRxiv}.

\bibitem[{Consortium et~al.(2021)Consortium, Bae, Baptiste, Bishop, Bodor, Brittain, Buchanan, Bumbarger, Castro, Celii et~al.}]{microns2021functional}
Consortium, M.; Bae, J.~A.; Baptiste, M.; Bishop, C.~A.; Bodor, A.~L.; Brittain, D.; Buchanan, J.; Bumbarger, D.~J.; Castro, M.~A.; Celii, B.; et~al. 2021.
\newblock Functional connectomics spanning multiple areas of mouse visual cortex.
\newblock \emph{BioRxiv}, 2021--07.

\bibitem[{Dorkenwald et~al.(2022)Dorkenwald, McKellar, Macrina, Kemnitz, Lee, Lu, Wu, Popovych, Mitchell, Nehoran et~al.}]{dorkenwald2022flywire}
Dorkenwald, S.; McKellar, C.~E.; Macrina, T.; Kemnitz, N.; Lee, K.; Lu, R.; Wu, J.; Popovych, S.; Mitchell, E.; Nehoran, B.; et~al. 2022.
\newblock FlyWire: online community for whole-brain connectomics.
\newblock \emph{Nature Methods}, 19(1): 119--128.

\bibitem[{Funke et~al.(2018)Funke, Tschopp, Grisaitis, Sheridan, Singh, Saalfeld, and Turaga}]{funke2018large}
Funke, J.; Tschopp, F.; Grisaitis, W.; Sheridan, A.; Singh, C.; Saalfeld, S.; and Turaga, S.~C. 2018.
\newblock Large scale image segmentation with structured loss based deep learning for connectome reconstruction.
\newblock \emph{IEEE transactions on pattern analysis and machine intelligence}, 41(7): 1669--1680.

\bibitem[{Hadsell, Chopra, and LeCun(2006)}]{hadsell2006dimensionality}
Hadsell, R.; Chopra, S.; and LeCun, Y. 2006.
\newblock Dimensionality reduction by learning an invariant mapping.
\newblock In \emph{Proceedings of the IEEE/CVF Conference on Computer Vision and Pattern Recognition (CVPR)}, volume~2, 1735--1742. IEEE.

\bibitem[{Hu, Shen, and Sun(2018)}]{hu2018squeeze}
Hu, J.; Shen, L.; and Sun, G. 2018.
\newblock Squeeze-and-excitation networks.
\newblock In \emph{Proceedings of the IEEE/CVF Conference on Computer Vision and Pattern Recognition (CVPR)}, 7132--7141.

\bibitem[{Januszewski et~al.(2018)Januszewski, Kornfeld, Li, Pope, Blakely, Lindsey, Maitin-Shepard, Tyka, Denk, and Jain}]{FFN}
Januszewski, M.; Kornfeld, J.; Li, P.~H.; Pope, A.; Blakely, T.; Lindsey, L.; Maitin-Shepard, J.; Tyka, M.; Denk, W.; and Jain, V. 2018.
\newblock High-precision automated reconstruction of neurons with flood-filling networks.
\newblock \emph{Nature methods}, 15(8): 605--610.

\bibitem[{Lee et~al.(2021)Lee, Lu, Luther, and Seung}]{lee2021learning}
Lee, K.; Lu, R.; Luther, K.; and Seung, H.~S. 2021.
\newblock Learning and segmenting dense voxel embeddings for 3D neuron reconstruction.
\newblock \emph{IEEE Transactions on Medical Imaging}, 40(12): 3801--3811.

\bibitem[{Lee et~al.(2017)Lee, Zung, Li, Jain, and Seung}]{lee2017superhuman}
Lee, K.; Zung, J.; Li, P.; Jain, V.; and Seung, H.~S. 2017.
\newblock Superhuman Accuracy on the {SNEMI3D} Connectomics Challenge.
\newblock \emph{arXiv preprint arXiv:1706.00120}.

\bibitem[{Li et~al.(2020)Li, Januszewski, Jain, and Li}]{VJain-MICCAI-2020}
Li, H.; Januszewski, M.; Jain, V.; and Li, P.~H. 2020.
\newblock Neuronal subcompartment classification and merge error correction.
\newblock In \emph{Medical Image Computing and Computer-Assisted Intervention}, 88--98.

\bibitem[{Li et~al.(2019)Li, Lindsey, Januszewski, Tyka, Maitin-Shepard, Blakely, and Jain}]{fafb-ffn}
Li, P.~H.; Lindsey, L.~F.; Januszewski, M.; Tyka, M.; Maitin-Shepard, J.; Blakely, T.; and Jain, V. 2019.
\newblock Automated reconstruction of a serial-section {EM} {D}rosophila brain with flood-filling networks and local realignment.
\newblock \emph{Microscopy and Microanalysis}, 25(S2): 1364--1365.

\bibitem[{Matejek et~al.(2019)Matejek, Haehn, Zhu, Wei, Parag, and Pfister}]{matejek2019biologically}
Matejek, B.; Haehn, D.; Zhu, H.; Wei, D.; Parag, T.; and Pfister, H. 2019.
\newblock Biologically-constrained graphs for global connectomics reconstruction.
\newblock In \emph{Proceedings of the IEEE/CVF Conference on Computer Vision and Pattern Recognition (CVPR)}, 2089--2098.

\bibitem[{Qi et~al.(2017)Qi, Yi, Su, and Guibas}]{PointNet++-NIPS-2017}
Qi, C.~R.; Yi, L.; Su, H.; and Guibas, L.~J. 2017.
\newblock Point{N}et++: Deep hierarchical feature learning on point sets in a metric space.
\newblock \emph{Neural Information Processing Systems}, 30.

\bibitem[{Saalfeld et~al.(2009)Saalfeld, Cardona, Hartenstein, and Toman{\v{c}}{\'a}k}]{saalfeld2009catmaid}
Saalfeld, S.; Cardona, A.; Hartenstein, V.; and Toman{\v{c}}{\'a}k, P. 2009.
\newblock CATMAID: collaborative annotation toolkit for massive amounts of image data.
\newblock \emph{Bioinformatics}, 25(15): 1984--1986.

\bibitem[{Sato et~al.(2000)Sato, Bitter, Bender, Kaufman, and Nakajima}]{sato2000teasar}
Sato, M.; Bitter, I.; Bender, M.~A.; Kaufman, A.~E.; and Nakajima, M. 2000.
\newblock TEASAR: tree-structure extraction algorithm for accurate and robust skeletons.
\newblock In \emph{Proceedings the Eighth Pacific Conference on Computer Graphics and Applications}, 281--449. IEEE.

\bibitem[{Schlegel, Gokaslan, and Kazimiers(2022)}]{philipp_schlegel_2022_7308283}
Schlegel, P.; Gokaslan, A.; and Kazimiers, T. 2022.
\newblock navis-org/skeletor: Version 1.2.3.

\bibitem[{Schroff, Kalenichenko, and Philbin(2015)}]{schroff2015facenet}
Schroff, F.; Kalenichenko, D.; and Philbin, J. 2015.
\newblock Facenet: A unified embedding for face recognition and clustering.
\newblock In \emph{Proceedings of the IEEE/CVF Conference on Computer Vision and Pattern Recognition (CVPR)}, 815--823.

\bibitem[{Shapson-Coe et~al.(2021)Shapson-Coe, Januszewski, Berger, Pope, Wu, Blakely, Schalek, Li, Wang, Maitin-Shepard, Karlupia, Dorkenwald, Sjostedt, Leavitt, Lee, Bailey, Fitzmaurice, Kar, Field, Wu, Wagner-Carena, Aley, Lau, Lin, Wei, Pfister, Peleg, Jain, and Lichtman}]{h01}
Shapson-Coe, A.; Januszewski, M.; Berger, D.~R.; Pope, A.; Wu, Y.; Blakely, T.; Schalek, R.~L.; Li, P.~H.; Wang, S.; Maitin-Shepard, J.; Karlupia, N.; Dorkenwald, S.; Sjostedt, E.; Leavitt, L.; Lee, D.; Bailey, L.; Fitzmaurice, A.; Kar, R.; Field, B.; Wu, H.; Wagner-Carena, J.; Aley, D.; Lau, J.; Lin, Z.; Wei, D.; Pfister, H.; Peleg, A.; Jain, V.; and Lichtman, J.~W. 2021.
\newblock A connectomic study of a petascale fragment of human cerebral cortex.
\newblock \emph{bioRxiv}.

\bibitem[{Sheridan et~al.(2022)Sheridan, Nguyen, Deb, Lee, Saalfeld, Turaga, Manor, and Funke}]{sheridan2022local}
Sheridan, A.; Nguyen, T.~M.; Deb, D.; Lee, W.-C.~A.; Saalfeld, S.; Turaga, S.~C.; Manor, U.; and Funke, J. 2022.
\newblock Local shape descriptors for neuron segmentation.
\newblock \emph{Nature Methods}, 1--9.

\bibitem[{Shi et~al.(2020)Shi, Yu, Liu, Zhang, and Li}]{shi2020optimal}
Shi, Y.; Yu, X.; Liu, L.; Zhang, T.; and Li, H. 2020.
\newblock Optimal feature transport for cross-view image geo-localization.
\newblock In \emph{Proceedings of the AAAI Conference on Artificial Intelligence}, volume~34.

\bibitem[{Wang et~al.(2021)Wang, Zhou, Yu, Dai, Konukoglu, and Van~Gool}]{Wang_2021_ICCV}
Wang, W.; Zhou, T.; Yu, F.; Dai, J.; Konukoglu, E.; and Van~Gool, L. 2021.
\newblock Exploring Cross-Image Pixel Contrast for Semantic Segmentation.
\newblock In \emph{Proceedings of the IEEE/CVF International Conference on Computer Vision (ICCV)}, 7303--7313.

\bibitem[{Wei et~al.(2021)Wei, Lee, Li, Lu, Bae, Liu, Zhang, dos Santos, Lin, Uram et~al.}]{wei2021axonem}
Wei, D.; Lee, K.; Li, H.; Lu, R.; Bae, J.~A.; Liu, Z.; Zhang, L.; dos Santos, M.; Lin, Z.; Uram, T.; et~al. 2021.
\newblock AxonEM dataset: 3d axon instance segmentation of brain cortical regions.
\newblock In \emph{Medical Image Computing and Computer-Assisted Intervention}, 175--185. Springer.

\bibitem[{Wolf et~al.(2018)Wolf, Pape, Bailoni, Rahaman, Kreshuk, Kothe, and Hamprecht}]{wolf2018mutex}
Wolf, S.; Pape, C.; Bailoni, A.; Rahaman, N.; Kreshuk, A.; Kothe, U.; and Hamprecht, F. 2018.
\newblock The mutex watershed: efficient, parameter-free image partitioning.
\newblock In \emph{Proceedings of the European Conference on Computer Vision (ECCV)}, 546--562.

\bibitem[{Yang et~al.(2018)Yang, She, Lai, and Yang}]{yang2018retrieving}
Yang, J.; She, D.; Lai, Y.-K.; and Yang, M.-H. 2018.
\newblock Retrieving and classifying affective images via deep metric learning.
\newblock In \emph{Proceedings of the AAAI Conference on Artificial Intelligence}, volume~32.

\bibitem[{Zheng et~al.(2018)Zheng, Lauritzen, Perlman, Robinson, Nichols, Milkie, Torrens, Price, Fisher, Sharifi et~al.}]{FAFB}
Zheng, Z.; Lauritzen, J.~S.; Perlman, E.; Robinson, C.~G.; Nichols, M.; Milkie, D.; Torrens, O.; Price, J.; Fisher, C.~B.; Sharifi, N.; et~al. 2018.
\newblock A complete electron microscopy volume of the brain of adult {D}rosophila melanogaster.
\newblock \emph{Cell}, 174(3): 730--743.

\bibitem[{Zhou et~al.(2022)Zhou, Wang, Konukoglu, and Van~Gool}]{zhou2022rethinking}
Zhou, T.; Wang, W.; Konukoglu, E.; and Van~Gool, L. 2022.
\newblock Rethinking semantic segmentation: A prototype view.
\newblock In \emph{Proceedings of the IEEE/CVF Conference on Computer Vision and Pattern Recognition (CVPR)}, 2582--2593.

\bibitem[{Zung et~al.(2017)Zung, Tartavull, Lee, and Seung}]{zung2017error}
Zung, J.; Tartavull, I.; Lee, K.; and Seung, H.~S. 2017.
\newblock An error detection and correction framework for connectomics.
\newblock \emph{Advances in neural information processing systems}, 30.

\end{thebibliography}

\end{document}